\documentclass[sigconf,natbib=true, nonacm]{acmart}
\usepackage{multirow}
\usepackage{xcolor}
\usepackage{bbm}
\usepackage[bottom]{footmisc}
\usepackage{float}
\usepackage{tikz}
\usepackage{xurl}
\usepackage{pgfplots}
\usepgfplotslibrary{groupplots}
\usepackage{graphics}
\usepackage{longtable}
\usepackage{caption}
\usepackage{hyperref}
\usepackage{subcaption}
\usepackage[most]{tcolorbox}
\pgfplotsset{compat=1.18}
\AtBeginDocument{%
  \providecommand\BibTeX{{%
    \normalfont B\kern-0.5em{\scshape i\kern-0.25em b}\kern-0.8em\TeX}}}

\begin{document}

%%
%% The "title" command has an optional parameter,
%% allowing the author to define a "short title" to be used in page headers.
\title{W-RAG: Weakly Supervised Dense Retrieval in RAG for Open-domain Question Answering}

%%
%% The "author" command and its associated commands are used to define
%% the authors and their affiliations.
%% Of note is the shared affiliation of the first two authors, and the
%% "authornote" and "authornotemark" commands
%% used to denote shared contribution to the research.

\author{Jinming Nian}
\affiliation{%
  \institution{Santa Clara University}
  % \streetaddress{500 El Camino Real}
  \city{Santa Clara}
  \state{CA}
  \country{USA}}
\email{jnian@scu.edu}

\author{Zhiyuan Peng}
\affiliation{%
  \institution{Santa Clara University}
  % \streetaddress{500 El Camino Real}
  \city{Santa Clara}
  \state{CA}
  \country{USA}}
\email{zpeng@scu.edu}

\author{Qifan Wang}
\affiliation{%
  \institution{Meta AI}
  % \streetaddress{1 Hacker Wy}
  \city{Menlo Park}
  \state{CA}
  \country{USA}}
\email{wqfcr@meta.com}

\author{Yi Fang}
%\authornote{Yi Fang is the corresponding author.}
\affiliation{%
  \institution{Santa Clara University}
  % \streetaddress{500 El Camino Real}
  \city{Santa Clara}
  \state{CA}
  \country{USA}}
\email{yfang@scu.edu}

%%
%% By default, the full list of authors will be used in the page
%% headers. Often, this list is too long, and will overlap
%% other information printed in the page headers. This command allows
%% the author to define a more concise list
%% of authors' names for this purpose.
% \renewcommand{\shortauthors}{Trovato and Tobin, et al.}

%%
%% The abstract is a short summary of the work to be presented in the
%% article.
\begin{abstract}

    In knowledge-intensive tasks such as open-domain question answering (OpenQA), large language models (LLMs) often struggle to generate factual answers, relying solely on their internal (parametric) knowledge. To address this limitation, Retrieval-Augmented Generation (RAG) systems enhance LLMs by retrieving relevant information from external sources, thereby positioning the retriever as a pivotal component. Although dense retrieval demonstrates state-of-the-art performance, its training poses challenges due to the scarcity of ground-truth evidence, largely attributed to the high costs of human annotation. In this paper, we propose W-RAG, a method that draws weak training signals from the downstream task (such as OpenQA) of an LLM, and fine-tunes the retriever to prioritize passages that most benefit the task. Specifically, we rerank the top-$k$ passages retrieved via BM25 by assessing the probability that the LLM will generate the correct answer for a question given each passage. The highest-ranking passages are then used as positive fine-tuning examples for dense retrieval. We conduct comprehensive experiments across four publicly available OpenQA datasets to demonstrate that our approach enhances both retrieval and OpenQA performance compared to baseline models, achieving results comparable to models fine-tuned with human-labeled data. Source code is published \footnote{\url{https://github.com/jmnian/WRAG}}. 

\end{abstract}

%%
%% The code below is generated by the tool at http://dl.acm.org/ccs.cfm.
%% Please copy and paste the code instead of the example below.
%%

% \begin{CCSXML}
% <ccs2012>
%    <concept>
%        <concept_id>10002951.10003317.10003347.10003348</concept_id>
%        <concept_desc>Information systems~Question answering</concept_desc>
%        <concept_significance>500</concept_significance>
%        </concept>
%  </ccs2012>
% \end{CCSXML}

% \ccsdesc[500]{Information systems~Question answering}

%%
%% Keywords. The author(s) should pick words that accurately describe
%% the work being presented. Separate the keywords with commas.
% \keywords{Retrieval Augmented Generation, Open-domain Question Answering, Weak Supervised Learning, Dense Retrieval}

%% A "teaser" image appears between the author and affiliation
%% information and the body of the document, and typically spans the
%% page.
% \begin{teaserfigure}
%   \includegraphics[width=\textwidth]{sampleteaser}
%   \caption{Seattle Mariners at Spring Training, 2010.}
%   \Description{Enjoying the baseball game from the third-base
%   seats. Ichiro Suzuki preparing to bat.}
%   \label{fig:teaser}
% \end{teaserfigure}

% \received{20 February 2007}
% \received[revised]{12 March 2009}
% \received[accepted]{5 June 2009}

%%
%% This command processes the author and affiliation and title
%% information and builds the first part of the formatted document.
\maketitle

\section{Introduction}\label{sec: intro}

Open-domain question answering (OpenQA) dating back to the 1960s \cite{DBLP:conf/aieeire/GreenWCL61} provides natural language-like answers to reply users' questions. OpenQA adopts the ``Retriever-Reader'' architecture \cite{DBLP:journals/corr/abs-2101-00774}, where the retriever retrieves relevant passages for the reader to generate answers. Previous studies \cite{rag, xiong2020answering} adopt a seq2seq model as a reader and train it on the labeled dataset. Recently, large language models (LLMs) like GPT-4 \cite{DBLP:journals/corr/abs-2303-08774} and LLaMA \cite{llama} have demonstrated astonishing performance on various tasks, attributed to the substantial amount of knowledge stored in their internal parameters. Despite the unprecedented achievements of LLMs, they face constraints such as the inability to consistently integrate up-to-date knowledge, as their parametric knowledge is fixed after being trained on huge datasets. Additionally, they are prone to generating plausible but non-factual responses, known as hallucinations \cite{DBLP:conf/iclr/WelleckKRDCW20}.

To overcome the limitations of LLMs' parametric knowledge, retrieval augmented generation (RAG) \cite{rag, DBLP:journals/corr/abs-2002-08909} is explored, equipping LLMs with a retriever to gather necessary evidence from external sources. Among the two components of RAG, improving the retriever is more feasible due to the recent trend of black-box APIs \cite{DBLP:journals/corr/abs-2303-08774} and the high cost and time requirements of fine-tuning open-source LLMs \cite{dubey2024llama}.

The retriever, a critical part of RAG, is typically either a traditional unsupervised retriever like BM25 \cite{bm25} or a more advanced neural retriever, such as dense retrieval \cite{monoT5, ance, DBLP:conf/sigir/KhattabZ20, DBLP:conf/emnlp/KarpukhinOMLWEC20}, which encodes questions and passages into the same embedding space and then measures the question-passage relevance score by vector similarity. A key challenge in training dense retrievers is the scarcity of human-annotated data, as in OpenQA, human-labeled evidence passages are often unavailable. Methods like UPR \cite{UPR}, which ranks passages based on the likelihood of LLMs generating the question given the passage, and AAR \cite{DBLP:conf/acl/YuXY023}, which combines the top-$k$ passages ranked by the LLM's averaged cross-attention scores with ground-truth passages as positive passages, have been employed to train dense retrievers. These approaches can train retrievers to find semantically relevant passages but do not guarantee improved RAG performance in OpenQA. As \citet{noise} demonstrated, retrieving relevant passages that cannot answer the question may negatively impact RAG performance in OpenQA tasks. Neither cross-attention scores nor the likelihood of LLMs generating the question based on the input passage explicitly ensures that the question can be answered by the “positive” passage. 

To address the scarcity of training data for dense retrievers in RAG for OpenQA, and to bridge the gap between semantic relevance and containing the actual answer, we propose extracting weak labels from existing OpenQA question-answer pairs by leveraging the ranking capabilities of LLMs \cite{rankgpt}. Specifically, we first use BM25 to retrieve the top-$k$ passages for a question, then pair each passage with the question. We rank the passages based on the likelihood that the LLM would generate the question’s ground-truth answer from each question-passage pair. Only the top-ranked passage is selected as the positive example for the question, and we train the dense retrievers using in-batch negative sampling. Our method evaluates the relevance of a question-passage pair by the likelihood of generating the correct answer for the question given the passage, making it particularly suitable for OpenQA, where retrieved passages must be capable of providing the correct answer. 

We conducted comprehensive experiments across four publicly available OpenQA datasets, and the results demonstrate that our approach enhances both retrieval and OpenQA performances compared to baseline models. Our contributions can be summarized as: 
\begin{itemize}
  \item We propose W-RAG, a general framework that uses answer likelihoods from an LLM given question-answer pairs to generate weak relevance labels for fine-tuning dense retrievers in RAG systems.  

  \item Comprehensive experiments demonstrate that W-RAG data improves both retrieval and OpenQA performance over baselines and achieves results comparable to models trained on human-labeled data.

  \item Open source LLMs were used, and the code was released to ensure reproducibility.
\end{itemize}

\section{Related Work}
\subsection{Dense Retrieval}
Traditional information retrieval (IR) methods are based on exact term matching, like BM25. While it is still widely used due to its efficiency, effectiveness, and robustness, it suffers from the well-known issue of lexical gap \cite{DBLP:conf/sigir/BergerCCFM00}. To address this, leveraging neural networks, dense retrieval (DR) employs pre-trained language models like BERT \cite{DBLP:conf/naacl/DevlinCLT19} to encode questions and passages into embeddings, and measures the question-passage relevance score by comparing the vectors similarly in the embedding space. Specifically, DR encodes the whole corpus into embeddings and builds the index, such as Faiss \cite{douze2024faiss}, on them. When a new question comes in, DR encodes the question into an embedding and performs a nearest neighbor search. DR can be classified into two categories: supervised, like DPR \cite{DBLP:conf/emnlp/KarpukhinOMLWEC20}, TAS-B \cite{DBLP:conf/sigir/HofstatterLYLH21}, and ColBERT \cite{DBLP:conf/sigir/KhattabZ20, DBLP:conf/naacl/SanthanamKSPZ22}; unsupervised, like Contriever\cite{DBLP:journals/tmlr/IzacardCHRBJG22} and ReContriever\cite{DBLP:conf/acl/LeiDCZYT23}, based on the training method. 

DPR utilizes a dual-tower architecture, with one BERT model dedicated to encoding questions and another to encoding passages, though in practice sometimes a single encoder is shared between passages and questions. The similarity between the question and passage embeddings is calculated, aiming to maximize the log-likelihood of the positive passage. ColBERT uses the same BERT model for both the question and passage encoders, differentiating them by appending a unique special token after the \texttt{[CLS]} token. Unlike DPR, which directly compares question and passage embeddings, ColBERT introduces a late interaction mechanism. It computes the similarity between each question token and all passage tokens, followed by maximum pooling over these similarities. The final similarity score for a question-passage pair is the sum of the pooled scores. TAS-B groups queries based on their embedding similarities and applies a training data sampling technique along with dual-teacher supervision distillation. Contriever trains a bi-encoder model using contrastive learning, generating positive question-passage pairs from an unlabeled corpus. ReContriever follows the same approach as Contriever for generating weak question-passage pairs but adds a self-scoring mechanism during training, where the loss is weighted by these scores.

\subsection{RAG for OpenQA}

RAG models have been applied to OpenQA, demonstrating significant performance improvement. Different RAG models have been proposed to solve the critical issues in OpenQA, such as how to retrieve relevant passages \cite{DBLP:conf/acl/YuXY023, DBLP:conf/naacl/ShiMYS0LZY24, DBLP:conf/emnlp/ShaoGSHDC23, DBLP:conf/acl/TrivediBKS23}, when to call the retriever \cite{DBLP:conf/emnlp/JiangXGSLDYCN23, DBLP:conf/emnlp/WangLSL23, DBLP:journals/corr/abs-2406-12534}, and how to decrease the computational complexity \cite{DBLP:conf/emnlp/0001DGL23, DBLP:journals/corr/abs-2310-06839, DBLP:conf/iclr/XuSC24, DBLP:conf/iclr/KimNMP0S0S24}. 

Since the retriever is a critical component in RAG, some studies have tried to improve the quality of retrieved passages, including training a better retriever \cite{DBLP:conf/acl/YuXY023, DBLP:conf/naacl/ShiMYS0LZY24} and prompt engineering \cite{DBLP:conf/emnlp/ShaoGSHDC23, DBLP:conf/acl/TrivediBKS23}. AAR \cite{DBLP:conf/acl/YuXY023} combines the top-$k$ passages ranked by the LLM's averaged cross-attention scores with ground-truth passages as positive passages and then follows ANCE \cite{DBLP:conf/iclr/XiongXLTLBAO21} to sample negative passages to train the dense retriever. ITER-RETGEN \cite{DBLP:conf/emnlp/ShaoGSHDC23} leverages the model output from the previous iteration as a specific context to help retrieve more relevant knowledge. IRCoT \cite{DBLP:conf/acl/TrivediBKS23} also adopts a similar approach to perform retrieval, but applies CoT for generating responses.

In some cases, LLMs can generate factual content well without external knowledge, and retrieving passages will decrease RAG's performance. Self-RAG \cite{DBLP:conf/emnlp/WangLSL23} and FLARE \cite{DBLP:conf/emnlp/JiangXGSLDYCN23} both fine-tune the LLMs to call the search engine automatically when external knowledge is needed. UAR \cite{DBLP:journals/corr/abs-2406-12534} trains classifiers to identify the need for external knowledge.

Applying all the top-$k$ retrieved passages as context not only increases the computational complexity and inference latency but also brings noise. Numerous methods have been proposed to compress the retrieved passages. Selective-Context \cite{DBLP:conf/emnlp/0001DGL23} filters non-essential lexical units by the summarization of self-information of each token contained in the unit. LongLLMLingua \cite{DBLP:journals/corr/abs-2310-06839} contrasts the perplexity score of each token in the passage with the perplexity score of the same token conditioned on the question and adopts this conservative perplexity score to filter out tokens. RECOMP \cite{DBLP:conf/iclr/XuSC24} leverages the summarization models to summarize the retrieved passages. SuRe \cite{DBLP:conf/iclr/KimNMP0S0S24} summarizes retrieved passages conditioned on each answer candidate generated by prompting the LLMs and then selects the top summarizations ranked by LLMs through a combination of pointwise and pairwise scoring methods.

\begin{figure*} [t] % t means top
  \centering
  \includegraphics[width=\linewidth]{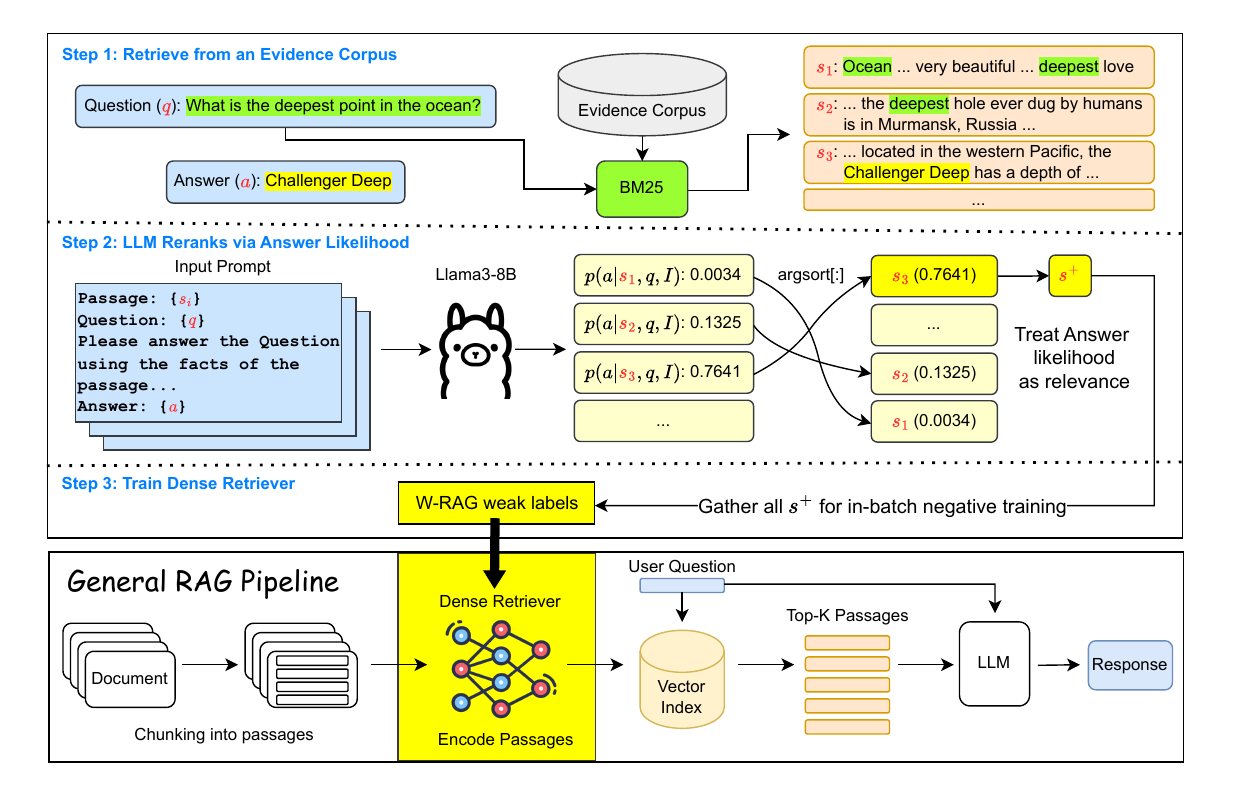}
  \caption{W-RAG fits into the general RAG pipeline by training the retriever with LLM-generated weak labels. Following the steps from top to bottom, we retrieve candidate passages using BM25, present each passage to the LLM to rerank based on the answer likelihood, then use the reranked top passage to train the retriever to finally enhance LLM's response quality through the standard RAG pipeline. Prompt in ``Step 2'' is shown in Figure~\ref{box:weak_prompt}.}
  \label{fig:main_diagram}
\end{figure*}

\subsection{LLM Signals for Retriever Training}

REPLUG \cite{DBLP:conf/naacl/ShiMYS0LZY24} improves dense retrieval by minimizing the KL divergence between the retrieval likelihood and LM likelihood of generating the second half of a text given the first half and a retrieved passage. The higher the likelihood that the LM will generate the second half correctly, the more relevant the retrieved passage is considered. Since only two separate corpora of text are needed to mine training signals, the method is highly scalable. CoRAG \cite{DBLP:journals/corr/abs-2501-14342} while focusing on multi-hop retrieval, selects training samples based on retrieval chains that maximize answer likelihood from the LLM. \cite{DBLP:journals/corr/abs-2406-12169} propose a teacher-student distillation method where they prompt the LLM to directly output ranked lists to fine-tune a reranker via list-wise ranking loss, while simultaneously aligning the retriever's output distribution with the reranker's. Similarly, ARL2 \cite{zhang-etal-2024-arl2} scores relevance by prompting LLMs to annotate passages as either supporting or not supporting the answer for a given question, then trains the retriever to align with these annotations.

\subsection{LLMs for Ranking}

Concerning the generation of weakly labeled ranking data using LLMs as passage rankers, existing work can be broadly grouped into three categories. The first category involves a zero-shot listwise ranking strategy, where a prompt containing the question, instructions, and multiple passages is given to the LLM, and the LLM generates a permutation of ranked indices \cite{twolar, listwise, rankvicuna, zero-shot-permutation}. The second category focuses on pointwise and pairwise relevance assessment. \cite{1-5} prompts the LLM with a passage and directly instructs it to rate its relevance, making it a point-wise relevance assessment method. \cite{pairwise} performs pair-wise ranking where two passages are given in the prompt and the LLM is asked to point out which is more relevant to the question. The third approach calculates the question generation likelihood when a passage is included in the prompt \cite{pointwise, query_prob1, query_prob2}. Our method for scoring passages aligns with this third approach. However, as mentioned in the introduction, our method ranks passages based on the likelihood that the LLM generates the ground-truth answer when conditioned on the question-passage pair. The ranked lists are then used as weakly labeled training data to fine-tune the retriever, with the goal of teaching it to retrieve passages that are most likely to condition the LLM to generate a correct answer. 

\begin{tcolorbox}[colframe=cyan!50!black, colback=cyan!11!white, title=Prompt for Answer Likelihood Extraction, width=\linewidth, boxrule=0.5mm, arc=2mm]

    Passage: \{passage\} \\
    Question: \{question\} \\
    Please answer the question using the facts of the passage. \\
    Keep your answer grounded to the facts of the passage. \\
    Keep your answer within one short sentence. \\ 
    Answer: \textbf{\{answer\}}

\end{tcolorbox}
\captionof{figure}{Bolded portion is evaluated for answer likelihood, which serves as a weak relevance score between ``passage'' and ``question''.}
\label{box:weak_prompt}

\section{Method}

As illustrated in Figure~\ref{fig:main_diagram}, W-RAG introduces a novel approach for fine-tuning the retrieval component within the RAG pipeline from a pre-trained dense retriever. Our method employs a weakly supervised training paradigm that requires only a set of question-answer pairs and a passage corpus. The W-RAG process unfolds in three stages: First, a first-stage retriever, such as BM25, retrieves candidate passages from the corpus. Second, we use an LLM to estimate the relevance of each passage by computing the likelihood of generating the correct answer when conditioned on the given question and passage, producing a weakly labeled ranking list. Finally, the dense retriever is fine-tuned using these weak labels. The resulting dense retriever is optimized for ranking passages with higher answer likelihood towards top positions, thereby enhancing the performance of LLMs across a wide array of tasks. 

\subsection{Weak-label Generation}

Given a passage corpus $\mathcal{C}=\{s_1, \dots, s_N\}$, we use a simple retriever to conduct first-stage retrieval. Given a question $q$, the retriever ranks each passage by relevance and selects the top-$k$ passages $\mathcal{P} = \{s'_1, \dots, s'_K\}$, where $\mathcal{P} \subset \mathcal{C}$. The passage corpus may comprise chunked Wikipedia articles or those from online sources, such as Reddit and Common Crawl, while the question-answer pairs could be sourced from QA datasets and question-answering platforms, such as Quora and Stack Overflow. The simple first-stage retriever could be lexical, such as BM25 \cite{bm25}, or a dense retriever. The primary objective at this stage is to maximize recall within a manageable set of retrieved passages. Ideally, this step fetches evidence passages containing the correct answer to the question. 

W-RAG is based on the hypothesis that an autoregressive LLM is more likely to generate the correct answer when the input passage contains the necessary information. We aim to leverage this signal by fine-tuning the retriever to rank passages according to their likelihood of eliciting the correct answer. Unlike traditional relevance measures, such as term overlap or semantic similarity, which do not emphasize that a passage enables answer generation. Our approach redefines relevance as the extent to which a passage conditions the LLM to produce the ground-truth answer. These labels serve as a weak supervision signal for fine-tuning the retriever.

As depicted in Figure~\ref{fig:main_diagram} Step 2, we construct weak label generation prompts with candidate passage $s'_i$, question $q$, some instructions $I$, and ground-truth answer $a$. The downstream signal is captured through the conditional probability of the ground-truth answer, denoted as $p(a|s'_i, q, I)$:  

\begin{equation}
    p(a|s'_i, q, I) = \prod_j \text{LLM}(a_j|s'_i, q, I, a_{<j})
\label{eq:likelihood}
\end{equation}

For each token in the input prompt, the LLM assigns logits, which, after applying a softmax operation, correspond to the probability of each vocabulary token being the next in the sequence. The probability of the $j$th answer token, denoted as $\text{LLM}(a_j|s'_i, q, I, a_{<j})$, is directly extracted from the logits of the preceding token. Since the ground-truth answer typically consists of multiple tokens, the cumulative probability can diminish rapidly. To address this, we take the logarithm on both sides of Equation~\ref{eq:likelihood}, converting each token's probability into log-likelihood. The overall probability of the answer $a$ is then computed as the average of the sum of the log-likelihoods for each token in the answer:

\begin{equation}
    \log p(a|s'_i, q, I) = \frac{1}{|a|} \sum_j \log \text{LLM}(a_j|s'_i, q, I, a_{<j})
\end{equation}

We consider this log-likelihood as the relevance score for the candidate passage $s'_i$. Each of the $k$ candidate passages is weakly labeled in this manner, resulting in a ranked list of passages, denoted as $\mathcal{S} = \{s'_{r_1}, \dots, s'_{r_K}\}$. This list is produced by sorting the set of candidate passages $\mathcal{S}$ according to their relevance scores.

\subsection{Training Dense Retriever}

Once we have accumulated a sufficient number of weakly labeled passages for a total of $M$ question-answer pairs, denoted as $\mathcal{W}=\{ \mathcal{S}_1 , \dots, \mathcal{S}_M\}$, we are ready to train a dense retriever that will assign higher scores to passages more likely to elicit the correct answer. In principle, any dense retriever, regardless of the specific objective function, can be trained using the generated weak labels. In this paper, we investigate two representative dense retrievers: DPR \cite{DPR} and ColBERT \cite{colbert}.

\subsubsection{\textbf{DPR}}

DPR utilizes a bi-encoder architecture, where the question and passage are independently mapped into a shared embedding space through two separate BERT encoders. In most practical implementations, a single shared BERT encoder is used for both the question and the passage, which is also the configuration used in our experiments. For a given question $q$ and passage $s$, the relevance score $R_{q, s}$ is given by the similarity between their respective \texttt{[CLS]} token embeddings, denoted as $\mathbf{e}_q \in \mathbb{R}^d$ and $\mathbf{e}_s \in \mathbb{R}^d$. We use cosine similarity as the similarity function:

\begin{equation}
    R_{q,s} = \cos (\mathbf{e}_{q}, \mathbf{e}_{s})
\end{equation}

Following the in-batch negative training method introduced in DPR, we process each ranked list $\mathcal{S} \in \mathcal{W}$ by extracting the top-ranked passage $s_i$ along with its associated question $q_i$. These pairs are then grouped into batches of size $n$ to form training batches, optimizing the retriever using a Multiple Negatives Ranking (MNR) loss \cite{MNR}. In this approach, for the $i$th question-passage pair, $s_i$ is treated as the positive example, while the other passages within the same batch $\{s_j\}_{j \ne i}$ are treated as negative examples. The MNR loss function is thus: 

\begin{equation}
    \mathcal{L}_{MNR} = - \sum_{i=1}^n \log (\frac{\exp (\alpha \cdot R_{q_i, s_i})}{\sum_{j=1}^n \exp (\alpha \cdot R_{q_i, s_j})})
\end{equation}

The scaler $\alpha$ amplifies the cosine similarity score, typically set to 20 according to the default setting in \texttt{sentence-transformers} \footnote{\url{https://sbert.net/docs/package_reference/sentence_transformer/losses.html\#multiplenegativesrankingloss}}.

\subsubsection{\textbf{ColBERT}}

% E is the full embedding. Say you have "[CLS] hi how are you [SEP]", the E will be a 768x6 matrix, whereas e is just the [CLS] token embedding which is 768x1 
ColBERT employs a bi-encoder architecture with late interaction, where the question and passage are independently encoded using a shared BERT model to obtain embeddings  $E_q \in \mathbb{R}^{l_q \times d}$ and $E_s \in \mathbb{R}^{l_s \times d}$, where $l_q$ and $l_s$ are the number of tokens in the question and passage, respectively, and $d$ is the embedding dimension. To compute the relevance score between a question and a passage, ColBERT applies a ``MaxSim'' operation: for each question token embedding $E_{q_i} \in E_q$, it finds the passage token embedding $E_{s_j} \in E_s$ that produces the highest dot product with $E_{q_i}$, and then sums these maximum values across all question tokens: 

\begin{equation} 
    R_{q,s} = \sum_{i=1}^{l_q} \max_{j \in [1, l_s]} \text{dot}(E_{q_i}, E_{s_j}) 
\label{colbert_sim_score} 
\end{equation}

Here, $\text{dot}(x,y)$ represents the dot product between vectors $x$ and $y$.
The training dataset for ColBERT consists of triplet samples $\langle q, s^+, s^- \rangle$, extracted from each ranked list $\mathcal{S}$. In each triplet, $s_{r_1}$ is treated as the positive passage $s^+$, while $s_{r_2}, s_{r_3}, \dots, s_{r_{m+1}}$ are treated as negative passages $s^-$. Here, $m$ represents the number of hard negatives selected from $\mathcal{S}$. ColBERT computes relevance scores $R_{q, s^+}$ and $R_{q, s^-}$ from each triplet, and optimizes the model using a pairwise softmax cross-entropy loss: 

\begin{equation}
    \mathcal{L}_{ColBERT} = -\log (\frac{\exp R_{q,s^+}}{\exp R_{q,s^+} + \exp R_{q,s^-}})
\end{equation}

\begin{table*}[t]
\centering
\resizebox{\linewidth}{!}{
\begin{tabular}{c|ccc|ccc|ccc|ccc}
\toprule
\multirow{2}{*}{ Retriever } & \multicolumn{3}{c}{ NQ } & \multicolumn{3}{c}{ SQuAD } & \multicolumn{3}{c}{ WebQ } & \multicolumn{3}{c}{ MSMARCO QnA } \\
\cmidrule(lr){2-4} \cmidrule(lr){5-7} \cmidrule(lr){8-10} \cmidrule(lr){11-13} 
& F1 & EM & BLEU-1 & F1 & EM & BLEU-1 & F1 & EM & BLEU-1 & F1 & EM & BLEU-1 \\

\midrule
\multicolumn{13}{c}{ Lower and upper bound of OpenQA performance } \\
\midrule
Naive            & 10.14 & 0.15  & 5.64  & 22.40 & 11.62 & 19.99 & 38.96 & 17.30 & 31.84 & 26.92 & 0.25 & 20.19 \\
Ground truth      & 69.95 & 54.90 & 65.43 & 77.69 & 58.10 & 72.99 & 52.39 & 33.76 & 46.08 & 43.40 & 9.95 & 28.44 \\
\midrule
\multicolumn{13}{c}{ Without fine-tuning } \\
\midrule
BM25             & 52.32 & 38.40 & 48.08 & 47.01 & 32.90 & 43.62 & 36.63 & 20.68 & 31.38 & 30.53 & 5.16 & 19.81 \\ 
Contriever       & 54.84 & 41.05 & 50.54 & 41.28 & 28.85 & 38.22 & 39.28 & 19.62 & 33.16 & 30.84 & 5.28 & 19.79 \\ 
ColBERT\textsubscript{init} & 17.95 & 11.60 & 15.91 & 16.82 & 10.90 & 15.27 & 20.94 & 12.24 & 18.13 & 19.24 & 1.80 & 12.79 \\
ReContriever     & 55.01 & 40.75 & 50.64 & 43.50 & 30.10 & 40.19 & 37.79 & 20.89 & 31.74 & 31.11 & \textbf{5.42} & 20.06 \\

\midrule
\multicolumn{13}{c}{ Fine-tuned with ground-truth data } \\
\midrule
ColBERT\textsubscript{base} & 53.95 & 39.85 & 49.65 & \textbf{52.43} & \textbf{37.37} & \textbf{48.77} & 34.21 & 19.20 & 29.35 & 30.51 & 4.60 & 19.79 \\
DPR\textsubscript{ReCon}    & \textbf{58.53} & \textbf{43.60} & \textbf{53.90} & 50.36 & 35.05 & 46.57 & \textbf{42.31} & \underline{23.42} & \textbf{35.91} & \underline{31.64} & \underline{5.32} & \underline{20.53} \\ 

\midrule
\multicolumn{13}{c}{ Fine-tuned with W-RAG data } \\
\midrule
ColBERT\textsubscript{base} & 50.35\textsuperscript{\textdagger} & 37.75\textsuperscript{\textdagger} & 46.42\textsuperscript{\textdagger} & \underline{52.28}\textsuperscript{\textdagger} & \underline{36.55}\textsuperscript{\textdagger} & \underline{48.32}\textsuperscript{\textdagger} & 32.16\textsuperscript{\textdagger} & 18.99\textsuperscript{\textdagger} & 27.69\textsuperscript{\textdagger} & 29.79\textsuperscript{\textdagger} & 4.95\textsuperscript{\textdagger} & 19.01\textsuperscript{\textdagger} \\
DPR\textsubscript{ReCon}    & \underline{56.40}\textsuperscript{\textdagger} & \underline{42.00}\textsuperscript{\textdagger} & \underline{51.98}\textsuperscript{\textdagger} & 49.71\textsuperscript{\textdagger} & 34.95\textsuperscript{\textdagger} & 45.99\textsuperscript{\textdagger} & \underline{41.59}\textsuperscript{\textdagger} & \textbf{23.63}\textsuperscript{\textdagger} & \underline{35.30}\textsuperscript{\textdagger} & \textbf{32.41}\textsuperscript{\textdagger} & 5.23 & \textbf{20.85} \\

\bottomrule
\end{tabular}
}
\caption{OpenQA performance using Llama3.1-8B-Instruct. The top 5 passages retrieved by each retriever are inserted into the QA prompt. \textbf{Bold} indicates the best performance; \underline{Underlined} indicates the second best (excluding ``Naive'' and ``Ground truth'')\textsuperscript{\textdagger}: Performance improvements are statistically significant with a t-test of $p < 0.05$. Prompt can be found in Figure~\ref{box:qa_prompt}.}
\label{tab: qa}
\end{table*}

\section{Experiments}

\subsection{Task and Datasets}

To assess the effectiveness of W-RAG generated weak labels, we study three crucial components: (1) the quality of the LLM-generated weak labels, (2) the retrieval performance of the weakly trained dense retriever, and (3) the end-to-end performance of the RAG system on the OpenQA task. We conduct experiments on four well-known datasets for OpenQA: MSMARCO QnA v2.1 \cite{nguyen2016ms}, NQ \cite{nq}, SQuAD \cite{squad}, and WebQ \cite{webq}. MSMARCO QnA v2.1 shares the same corpus as MSMARCOv1 Passage Retrieval, with questions originating from real user queries submitted to Bing. The language is generally conversational, with answers that are typically longer. NQ also features real user questions and uses a corpus of documents from the English Wikipedia. The answers are often very short and clean. In this work, we use a chunked version of the NQ corpus as prepared by Karpukhin et al. \cite{DPR}. SQuAD's corpus is similarly derived from the English Wikipedia, where the passages were first retrieved and then sampled. The questions and answers are manually written by crowdworkers, so they do not reflect natural language as closely as those from real user queries, such as MSMARCO QnA and NQ, and the answers are typically concise and focused. WebQ's corpus is drawn from Freebase, a large knowledge graph. Due to the nature of knowledge graphs, WebQ's questions and answers are entity-related and factoid-based, making them less conversational. Similar to NQ and SQuAD, the answers in WebQ are generally short.  

For each dataset, we uniformly random sampled 5,000 question-answer pairs and a corpus of 500,000 passages from the training set, ensuring that all questions had relevant passages included in the corpus. We then split the question-answer pairs into 2,000 as training set, 1,000 as validation set, and 2,000 as test set. This sampling was necessitated by resource and time constraints. We argue that this small training set is sufficient to demonstrate statistically significant improvements in both retrieval and the final OpenQA performance. While our results are indicative, our method could benefit from further validation using more weakly labeled training data to ensure generalizability.

\subsubsection{\textbf{Weak Label Quality}}

For our main experiments, we use Llama3-8B-Instruct \cite{llama3} to serve as the weak supervision labeler. The 2,000 question-answer pairs in the training set are used to generate weak labels. For each question, we score the top 100 passages retrieved by BM25 based on the LLM's likelihood of generating the ground-truth answer. We chose BM25 for its effectiveness and efficiency, though in principle this first-stage retrieval can be done with any retriever. The labeling process requires the LLM to perform 100 inference calls using the prompt shown in Figure~\ref{box:weak_prompt} to produce a ranked list. The choice of 100 passages is a trade-off between accuracy and latency. We found that BM25's Recall@100 reaches approximately 80\% across all four datasets, and retrieving additional passages yields diminishing returns as the LLM's inference time increases linearly. We evaluate the reranking performance using different LLMs, which will be elaborated on later in the ablation studies (\ref{sec:ablations}).

\begin{tcolorbox}[colframe=orange!50!black, colback=orange!5!white, title=Question Answering Prompt, width=\linewidth, boxrule=0.5mm, arc=2mm]
    \textbf{[System Prompt]}
    
    You are a helpful assistant who answers questions using the provided passage. If the passage does not contain the necessary information, you will answer the question directly based on your general knowledge. Only provide the final answer, no explanations.
    
    \vspace{0.2cm}
    
    \textbf{[User Prompt]}
    
    Passage: \{passage \#1\} \\
    Passage: \{passage \#2\} \\
    Passage: \{passage \#3\} \\ 
    Passage: \{passage \#4\} \\ 
    Passage: \{passage \#5\} \\ 
    Question: \{question\}

    \vspace{0.2cm}
    
    \textbf{[Assistant]}
\end{tcolorbox}
\captionof{figure}{Prompt for RAG OpenQA where top 5 retrieved passages are used.}
\label{box:qa_prompt}

\subsubsection{\textbf{Weakly Trained Retriever}}

After the LLM weakly labels the training set, we use these labels to train DPR and ColBERT. For DPR, since we use in-batch negatives during fine-tuning, we only need to select the top-ranked passage to curate a list of relevant question-passage pairs, thus forming 2,000 training samples. For all four datasets, we experiment with two different initialization settings for DPR. The first uses \texttt{bert-base-uncased} as the starting point and fine-tunes with W-RAG data from the pre-trained base BERT model (denoted as ``DPR\textsubscript{base}''). The second initializes from an unsupervised pre-trained dense retriever, \texttt{Yibin-Lei/ReContriever} \cite{recontriever}, and fine-tunes it with W-RAG data (denoted as ``DPR\textsubscript{ReCon}''). 

ColBERT's training process differs slightly from that of DPR, as it requires sampling of negative passages. For each question, we select the top-ranked passage from the reranked list as the positive passage and the immediate following 10 passages as hard negatives. According to the definition of our weak labels, these hard negatives are relevant but not sufficiently informative to elicit the correct answer from the LLM. We did not explore varying the number of hard negatives, as this is not the primary focus of this work. Additionally, since the off-the-shelf ColBERTv2 is trained on the entire MSMARCO dataset, using it as the starting point would introduce data leakage. For consistency, we always fine-tune ColBERT from the pre-trained base BERT model \texttt{bert-base-uncased} for all four datasets (denoted as ``ColBERT\textsubscript{base}'').

\subsubsection{\textbf{OpenQA Performance}}

Once the dense retriever is trained, we integrate it into the generic RAG pipeline, where the retriever is used to index the evidence corpus and retrieve the top-$k$ evidence passages for a given user question. The top 5 retrieved passages are directly inserted into the prompt, as shown in Figure~\ref{box:qa_prompt}. Llama3.1-8B-Instruct is used to perform OpenQA, and we restrict the maximum output token to be 20 tokens, following \cite{DBLP:journals/corr/abs-2405-13576}. We also explored the impact of using different numbers of supplemental evidence passages on both OpenQA performance and latency, which we discuss in the ablation studies (\ref{sec:ablations}). We did not experiment with different LLMs for the OpenQA task, as the LLM component of RAG is not the primary focus of our study.

\begin{figure*} % t means top, h means here
  \centering
  \includegraphics[width=\linewidth]{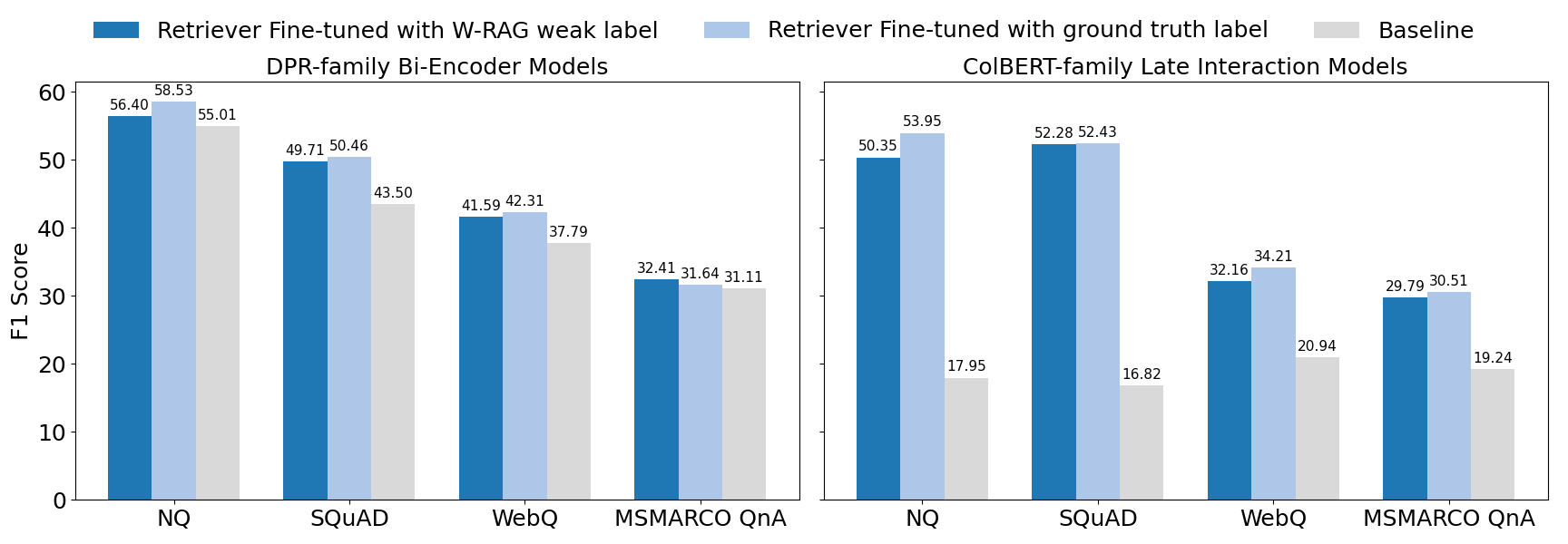}
  \caption{Same information as Table~\ref{tab: qa}, comparing the impact of retrievers fine-tuned with weak labels versus ground-truth labels on OpenQA performance, both of which are fine-tuned from the baseline model. ReContriever serves as the baseline for the DPR-family; ColBERT initialized from base BERT serves as the baseline for the ColBERT-family. } 
  \label{fig:qa_compare}
\end{figure*}

\subsection{Baselines}

Since our retriever is fine-tuned with only limited data, we do not compare it to state-of-the-art retrievers or RAG systems. Instead, we focus on the improvement from starting point baselines to our weakly supervised models and compare them with models fine-tuned on ground-truth data. For OpenQA evaluation, we use Llama3.1-8B-Instruct as the LLM in the RAG pipeline, assessing the impact of the final OpenQA performance using top passages retrieved by different retrievers. Additionally, we introduce two baselines: ``Naive,'' where Llama3 answers the question without any external information, showcasing its parameter knowledge, and ``Ground truth,'' where the ground-truth passage is inserted into the prompt, representing the best possible OpenQA performance in our setup. 

To assess the quality of weak labels, we compare the weakly labeled ranking lists with BM25 on the training set of each dataset. For the fine-tuned retriever's retrieval performance, we evaluate three categories of models. The first category includes entirely unsupervised models: BM25, ColBERT initialized with \texttt{bert-base-uncased} (denoted as ``ColBERT\textsubscript{init}''), Contriever \cite{contriever}, and ReContriever \cite{recontriever}. Contriever learns by contrasting sampled passages from different documents, assuming passages sampled from the same document are more similar than those sampled from different documents. ReContriever enhances Contriever by adding an estimated prior of the relevance among sampled passages. The second category includes retrievers fine-tuned on ground-truth data, and the third category are fine-tuned with W-RAG data. For a fair comparison between the second and third categories, we use the same questions and an equal number of positive and negative samples as training data. 

\subsection{Experimental Settings}

We begin by retrieving the top 100 passages using BM25 with the \texttt{rank\_bm25} \footnote{\url{https://github.com/dorianbrown/rank_bm25}} BM25Okapi with $k_1$=1.5, $b$=0.75 and $\epsilon$=0.25. The \texttt{nltk} \footnote{\url{https://github.com/nltk/nltk}} tokenizer is used to tokenize the questions and passages. The prompt used to generate weak labels is shown in Figure~\ref{box:weak_prompt}. The weak relevance score between a question and a passage is extracted from the logits of the answer tokens, which is why the ground-truth answer is explicitly included in the prompt. We use BM25's retrieval performance as a baseline and try different orderings of the passage, question, and instructions, finding that the weak labels generated from the ``Passage-Question-Instruction'' format consistently yield better retrieval metrics compared to the ``Passage-Instruction-Question'' and ``Instruction-Passage-Question'' formats. For fine-tuning, we borrow BEIR's code \footnote{\url{https://github.com/beir-cellar/beir}} to manage the training data. The DPR models are fine-tuned using \texttt{sentence-transformers} with the following hyperparameters: a batch size of 128, a learning rate of $2\times 10^{-5}$, the AdamW optimizer, 20 epochs, and model weights are saved based on the best Recall@5 on the validation set. ColBERT is fine-tuned from base BERT models using RAGatouille \footnote{\url{https://github.com/bclavie/RAGatouille}} with a batch size of 64, 10 hard negatives per positive, learning rate of $1\times 10^{-5}$, AdamW optimizer, and one epoch. Generating weak labels took approximately two days per dataset on a single Nvidia V100 GPU. Our experiments are easily reproducible since BM25, ColBERT, DPR, Llama3-8B-Instruct, Llama3.1-8B-Instruct, and all four datasets are publicly accessible. 

\subsection{Evaluation}

To evaluate retrieval performance, we use Recall as our primary metric, as it reflects whether the relevant passage is included within a specific top-$k$ ranking range. We also report Mean Reciprocal Rank (MRR), which measures the inverse of the rank at which the first relevant passage appears. Unlike Recall, which only considers whether a relevant passage is retrieved, MRR also takes into account its position in the ranked list.

For OpenQA metrics, we use F1, Exact Match (EM), and BLEU-1 \cite{bleu} to assess the alignment between LLM-generated answers and ground-truth answers. F1 measures the token-level overlap between generated and reference answers. EM evaluates whether the generated answer exactly matches any of the ground-truth answers. BLEU-1 computes unigram precision, measuring the proportion of overlapping words between the LLM-generated answer and the reference answers without considering word order or recall. We evaluate OpenQA performance on our randomly sampled test set, which contains 2,000 questions per dataset. For questions with multiple ground-truth answers, the generated answer is compared with all references, and the highest resulting score is used for evaluation.

\begin{table*}[t]
\centering
\resizebox{\linewidth}{!}{
\begin{tabular}{c|ccc|ccc|ccc|ccc}
\toprule
\multirow{2}{*}{ Retriever } & \multicolumn{3}{c}{ NQ } & \multicolumn{3}{c}{ SQuAD } & \multicolumn{3}{c}{ WebQ } & \multicolumn{3}{c}{ MSMARCO QnA } \\
\cmidrule(lr){2-4} \cmidrule(lr){5-7} \cmidrule(lr){8-10} \cmidrule(lr){11-13}
& MRR@5 & R@1 & R@5 & MRR@5 & R@1 & R@5 & MRR@5 & R@1 & R@5 & MRR@5 & R@1 & R@5 \\
\midrule
\multicolumn{13}{c}{ Without fine-tuning } \\
\midrule

BM25 & 56.86 & 8.69 & 35.75 & \textbf{29.78} & 4.00 & \underline{20.11} & 27.37 & 2.85 & 13.36 & 27.81 & 16.47 & 45.80 \\
Contriever & 55.24 & 9.88 & 35.85 & 22.21 & 2.94 & 16.06 & 29.90 & 1.65 & 13.04 & 28.26 & 15.85 & 49.01 \\ 
ColBERT\textsubscript{init} & 8.66 & 0.93 & 2.24 & 5.04 & 0.63 & 1.58 & 0.31 & 0.01 & 0.04 & 1.63 & 1.20 & 2.20 \\
ReContriever & 58.31 & 11.18 & 39.52 & 21.50 & 2.84 & 15.97 & 30.22 & 1.94 & 13.70 & 29.08 & 15.89 & 50.93 \\
\midrule
\multicolumn{13}{c}{ Fine-tuned with ground-truth data } \\
\midrule
DPR\textsubscript{base} & 56.08 & 11.74 & 36.19 & 17.86 & 1.99 & 10.54 & \underline{40.28} & \underline{5.61} & \underline{20.59} & 32.37 & \underline{21.17} & 49.94 \\
ColBERT\textsubscript{base} & 59.25 & \underline{13.16} & 30.82 & 23.10 & \textbf{4.69} & 10.72 & 11.57 & 1.80 & 5.01 & 34.84 & 20.97 & 51.12 \\
DPR\textsubscript{ReCon} & \textbf{66.32} & \textbf{15.16} & \textbf{47.15} & \underline{28.29} & 3.93 & \textbf{20.72} & \textbf{41.55} & \textbf{6.40} & \textbf{22.62} & \textbf{36.52} & \textbf{22.84} & \textbf{59.13} \\

\midrule
\multicolumn{13}{c}{ Fine-tuned with W-RAG data } \\
\midrule
DPR\textsubscript{base} & 53.02 & 10.12 & 33.64 & 17.81 & 2.10 & 10.20 & 33.77 & 4.78 & 16.56 & 29.81 & 18.09 & 48.13 \\ 
ColBERT\textsubscript{base} & 52.29\textsuperscript{\textdagger} & 9.82\textsuperscript{\textdagger} & 26.72\textsuperscript{\textdagger} & 22.65\textsuperscript{\textdagger} & \underline{4.45}\textsuperscript{\textdagger} & 10.29\textsuperscript{\textdagger} & 8.31\textsuperscript{\textdagger} & 1.53\textsuperscript{\textdagger} & 3.26\textsuperscript{\textdagger} & 31.16\textsuperscript{\textdagger} & 19.73\textsuperscript{\textdagger} & 49.04\textsuperscript{\textdagger} \\
DPR\textsubscript{ReCon} & \underline{63.01}\textsuperscript{\textdagger} & 13.11\textsuperscript{\textdagger} & \underline{42.67}\textsuperscript{\textdagger} & 28.11\textsuperscript{\textdagger} & 3.82 & 20.01\textsuperscript{\textdagger} & 36.10\textsuperscript{\textdagger} & 4.99\textsuperscript{\textdagger} & 18.80\textsuperscript{\textdagger} & \underline{34.96}\textsuperscript{\textdagger} & 20.23\textsuperscript{\textdagger} & \underline{58.87}\textsuperscript{\textdagger} \\

\bottomrule
\end{tabular}
}
\caption{Retrieval Performance after training with W-RAG or human-labeled data. \textbf{Bold} indicates the best performance; \underline{Underlined} indicates the second best. \textsuperscript{\textdagger}: Performance improvements are statistically significant with a t-test of $p<0.05$. }
\label{tab: retrieval}
\end{table*}

\begin{table*}[t]
\centering
\resizebox{\linewidth}{!}{
\begin{tabular}{c|ccc|ccc|ccc|ccc}
\toprule
\multirow{2}{*}{Retriever} & \multicolumn{3}{c}{NQ} & \multicolumn{3}{c}{SQuAD} & \multicolumn{3}{c}{WebQ} & \multicolumn{3}{c}{MSMARCO QnA} \\
\cmidrule(lr){2-4} \cmidrule(lr){5-7} \cmidrule(lr){8-10} \cmidrule(lr){11-13}
 & R@1 & R@5 & R@50 & R@1 & R@5 & R@50 & R@1 & R@5 & R@50 & R@1 & R@5 & R@50 \\
\midrule
BM25 & 8.8 & 35.3 & 79.41 & 4.49 & 20.69 & 64.61 & 2.17 & 11.55 & 57.65 & 16.94 & 44.64 & 75.77 \\
Llama3-8B-Instruct & 15.42 & 42.24 & 73.03 & 8.05 & 30.88 & 67.74 & 4.94 & 18.87 & 55.02 & 52.39 & 66.14 & 75.17 \\
\bottomrule
\end{tabular}
}
\caption{BM25 retrieves the top 100 passages, which are then reranked by Llama3-8B-Instruct. The reranked lists are used as weak labels for training a dense retriever.}
\label{tab:weak_label_quality}
\end{table*}

\section{Experimental Results}

In this section, we discuss the final OpenQA performance, the retrieval performance of fine-tuned retrievers, the quality of weak labels, and the results of ablation studies. 

\subsection{Main Results}

The goal of W-RAG is to draw weak training signals from the downstream task (such as OpenQA) of an LLM, and fine-tune the retriever to prioritize passages that most benefit the task. In other words, W-RAG aims to improve OpenQA performance by aligning retrieval with answer generation. We highlight the performance improvements of weakly supervised retrievers compared to baselines, and demonstrate that the gap between weakly supervised and ground-truth supervised retrievers is relatively small. This trend is observed in both retrieval metrics and final performance in the RAG-based OpenQA setting. Table~\ref{tab: qa} presents the main OpenQA results using different retrievers in the RAG pipeline, evaluated on 2,000 test questions for each dataset. For these experiments, the top 5 retrieved passages are added to the prompt, as shown in Figure~\ref{box:qa_prompt}. The ``Naive'' and ``Ground truth'' represent the lower and upper bounds of OpenQA performance under our setup, respectively. The ``Naive'' baseline excludes any supplementary passage, while ``Ground truth'' includes the ground-truth passage. Any retriever that retrieves relevant passages should outperform the ``Naive'' baseline, while surpassing the ``Ground truth'' upper bound would be very unexpected. Figure~\ref{fig:qa_compare} provides a more intuitive comparison of the OpenQA results. Across all datasets, W-RAG-tuned retrievers consistently outperform baselines within both the DPR and ColBERT model families. Most of these improvements are statistically significant. Furthermore, the relatively small performance gap between W-RAG-tuned and ground-truth-tuned retrievers suggests that the weak labels generated by W-RAG approach the quality of human-labeled data. 

One baseline, ColBERT\textsubscript{init} (initialized with \texttt{bert-base-uncased}), underperforms the ``Naive'' baseline, likely due to irrelevant passages being retrieved, introducing noise that misguides the LLM. For MSMARCO QnA, the EM values are relatively smaller because the ground-truth answers are generally much longer than NQ, SQuAD, or WebQ, and are therefore more challenging for the LLM to generate exactly the same words as the ground-truth answer. 

\subsection{Retrieval Results}

The retrieval results of different fine-tuned retrievers are presented in Table~\ref{tab: retrieval}. These results are based on 2,000 test questions for each fine-tuned or baseline retriever. On NQ, WebQ, and MSMARCO QnA datasets, all retrievers fine-tuned on W-RAG data outperformed their corresponding baselines, while DPR\textsubscript{ReCon} consistently outperformed all unsupervised baselines, except for SQuAD where BM25 performed best. As expected, DPR fine-tuned from ReContriever showed better retrieval performance than DPR\textsubscript{base}, which was fine-tuned from the base BERT model. Similar to the OpenQA task, retrievers fine-tuned with W-RAG data slightly underperform those fine-tuned with ground-truth data. 

Although there is a general positive correlation between retrieval performance and OpenQA performance, the relationship is noisy and sometimes inconsistent. For instance, on SQuAD, BM25 achieved a much higher Recall@5 (20.11) compared to ColBERT\textsubscript{base} fine-tuned with W-RAG data (10.29), meaning BM25 ranked the relevant passage among the top 5 for approximately 200 more questions. However, in OpenQA, using the top 5 passages from ColBERT\textsubscript{base} led to better answer accuracy than using those from BM25. Similarly in NQ, DPR\textsubscript{ReCon} fine-tuned on W-RAG achieved significantly higher retrieval performance than ColBERT\textsubscript{base} (Recall@5 of 42.67 vs. 26.72), their OpenQA performances were relatively close.

This discrepancy may be partly due to the prompt instructing Llama3.1-8B-Instruct to fall back on its internal knowledge when the retrieved passage cannot answer the question. Nevertheless, these findings suggest that traditional notions of relevance may not directly translate into improved quality of answers in RAG systems. Retrievers trained on W-RAG data select top passages based on their ability to elicit the correct answer from the LLM, while ground-truth passages are typically chosen based on term overlap, semantic similarity, and human judgment. Contrary to expectations, stronger retrieval metrics do not always lead to better OpenQA results. This raises a critical question: \textit{Is the relevance definition in existing datasets truly effective for evaluating retrievers in RAG systems?}

\subsection{W-RAG Labels}

After generating the W-RAG data, we evaluate its retrieval performance and assess the quality of the weakly labeled ranking lists. As shown in Table~\ref{tab:weak_label_quality}, there is a significant gap in Recall@1 between BM25 and the reranking performance of Llama3-8B-Instruct. This observation supports our decision to use only the top-1 passage as the positive example when fine-tuning different retrievers. We also observe that the quality of the reranked list deteriorates at higher recall values. This decline occurs because the difference in answer likelihood between adjacent passages becomes negligible beyond a certain rank. For example, the difference in answer likelihood between the 50th and 51st ranked passages is approximately $\exp(-3.24815) - \exp(-3.24899) \approx 3\times 10^{-5}$, which is very insignificant. For passages that do not meaningfully increase the LLM's likelihood of generating the correct answer, the assigned likelihood scores remain similarly low across the ranking. 

\begin{figure} [t] % t means top, h means here
  \centering
  \includegraphics[width=\linewidth]{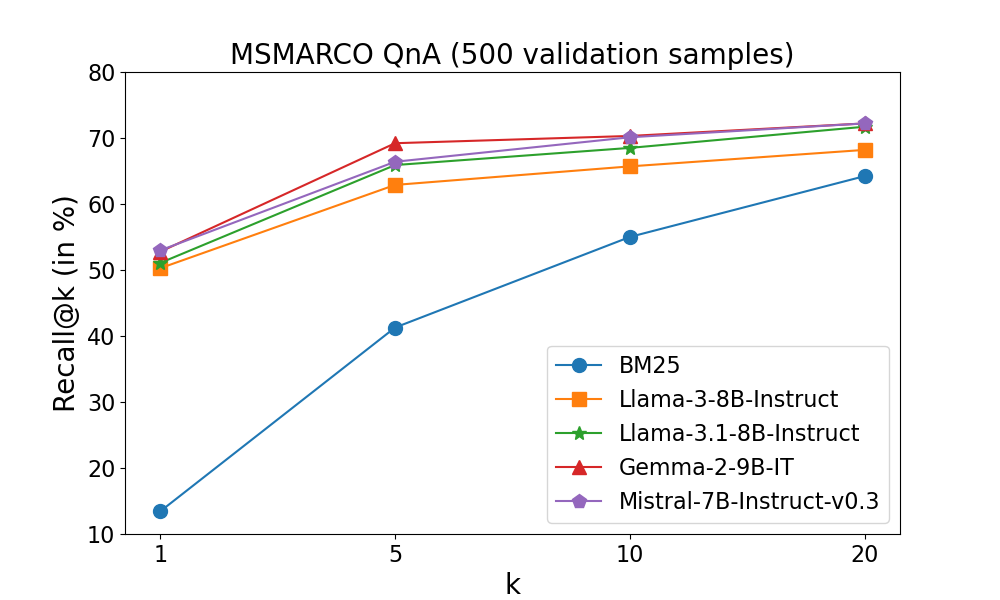}
  \caption{Comparison of recall for various LLMs at different top k positions, when reranking top 100 passages retrieved by BM25. } 
  \label{fig:diff_llm}
\end{figure}

\subsection{Ablation Study}
\label{sec:ablations}

To study the quality of W-RAG data as generated by different LLMs, we experimented with several LLMs \cite{DBLP:journals/corr/abs-2310-06825, DBLP:journals/corr/abs-2408-00118}, as shown in Figure~\ref{fig:diff_llm}. The figure reports reranking performance on 500 questions randomly selected from the validation set of MSMARCO QnA. Although the LLMs differ slightly in Recall@5, they exhibit consistent trends. This consistency gives us confidence that the W-RAG method generalizes well to models at the 8B parameter scale. We leave the more extensive study with larger models for future work. We notice Llama3-8B-Instruct is relatively weaker when compared to other examined LLMs. Nevertheless, the recall distributions across LLMs are similar, and we conclude that any of these models can effectively serve as the agent for generating weak supervision signals. In the 8B parameter scale, this choice does not substantially impact the outcome of the trained retrievers and the downstream OpenQA performance. 

We also examine the trade-off between OpenQA accuracy and latency on the NQ dataset by varying the number of top passages retrieved by DPR\textsubscript{ReCon}, as shown in Table~\ref{tab:num_docs}. We observe a consistent and steady improvement in all OpenQA metrics as more evidence passages are included in the prompt. This trend begins to plateau between 10 and 20 passages under our experimental setting. This behavior is expected: supplying more passages increases the likelihood that at least one of them contains the correct answer. However, latency increases linearly with the number of inserted passages, making it a trade-off between answer quality and inference speed.

\begin{table}[t]
\centering
% \resizebox{\linewidth}{!}{
\begin{tabular}{c|cccc}
\toprule
\# Passages & F1 & EM & BLEU-1 & Avg Latency (ms)\\
\midrule
1 & 54.64 & 40.95 & 50.50 & 264.90 \\
3 & 57.74 & 42.90 & 53.14 & 342.01 \\
5 & 60.21 & 45.35 & 55.62 & 413.96\\
10 & 61.32 & 46.10 & 56.59 & 602.33 \\ 
20 & 61.39 & 45.40 & 56.15 & 1005.50 \\
\bottomrule
\end{tabular}
% }
\caption{OpenQA performance on NQ using different numbers of top passages retrieved by DPR\textsubscript{ReCon} fine-tuned with W-RAG data. }
\label{tab:num_docs}
\end{table}

\section{Conclusions and Future Work}

This paper introduces a general framework, W-RAG, for extracting weak signals from question-answer pairs using an LLM to address the scarcity of training data for dense retrieval in RAG for OpenQA. W-RAG reranks the top-$k$ initially retrieved passages by the probability of LLMs generating the ground-truth answer conditioned on each question-passage pair. This ranking score computation method aligns well with a recent study \cite{noise}, which states that the retrieved passage should answer the question; otherwise, performance will be negatively impacted. Extensive experimental results on four public OpenQA datasets demonstrate the effectiveness of W-RAG. For future work, we plan to scale the generation of W-RAG data to a quantity comparable to that used in training state-of-the-art retrievers, and evaluate whether retrievers trained with W-RAG can surpass them in final OpenQA performance when integrated into a RAG system. As for future research directions, we plan to explore which types of passages most effectively enhance RAG's performance in OpenQA, as indicated by \cite{noise}, where even randomly sampled tokens were beneficial in some cases. Understanding the types of passages preferred by LLMs will allow us to design more effective dense retrieval methods, including new structures and evaluation metrics. Additionally, the compression of retrieved passages warrants further study, as directly feeding all retrieved passages to LLMs not only increases computational complexity but also introduces significant noise. Finally, even with a ground-truth evidence passage, RAG can still produce incorrect answers, known as hallucinations; enhancing the robustness of RAG in OpenQA is another promising direction for future research. 

% \clearpage
% \newpage
\balance
\bibliographystyle{ACM-Reference-Format}
\bibliography{arxiv}

\end{document}